\documentclass[conference]{IEEEtran}
\usepackage{amsmath,amsfonts}
\usepackage{multirow}

\usepackage{array}

\usepackage[caption=false,font=normalsize,labelfont=sf,textfont=sf]{subfig}
\usepackage{hyperref} 
\usepackage{textcomp}
\usepackage[ruled,vlined]{algorithm2e}
\usepackage{stfloats}
\usepackage{url}
\def\BibTeX{{\rm B\kern-.05em{\sc i\kern-.025em b}\kern-.08em
    T\kern-.1667em\lower.7ex\hbox{E}\kern-.125emX}}
\usepackage{verbatim}
\usepackage{graphicx}
\usepackage{cite}
\usepackage{graphicx}
\usepackage{subcaption}
\begin{document}

\title{\vspace{0.2in} QuickGrasp: Lightweight Antipodal Grasp Planning with Point Clouds}

\author{\IEEEauthorblockN{Navin~Sriram~Ravie*, Keerthi~Vasan~M*, Asokan~Thondiyath, Bijo~Sebastian}
\IEEEauthorblockA{\textit{Department of Engineering Design} \\
\textit{Indian Institute of Technology, Madras }\\
Chennai, India \\
bijo.sebastian@iitm.ac.in,
* Equal contribution}
}


\markboth{IEEE ROBOTICS AND AUTOMATION LETTERS }%
{Shell \MakeLowercase{\textit{et al.}}: A Sample Article Using IEEEtran.cls for IEEE Journals}


\maketitle

\begin{abstract}
Grasping has been a long-standing challenge in facilitating the final interface between a robot and the environment. As environments and tasks become complicated, the need to embed higher intelligence to infer from the surroundings and act on them has become necessary. Although most methods utilize techniques to estimate grasp pose by treating the problem via pure sampling-based approaches in the six-degree-of-freedom space or as a learning problem, they usually fail in real-life settings owing to poor generalization across domains. In addition, the time taken to generate the grasp plan and the lack of repeatability, owing to sampling inefficiency and the probabilistic nature of existing grasp planning approaches, severely limits their application in real-world tasks. This paper presents a lightweight analytical approach towards robotic grasp planning, particularly antipodal grasps, with little to no sampling in the six-degree-of-freedom space. The proposed grasp planning algorithm is formulated as an optimization problem towards estimating grasp points on the object surface instead of directly estimating the end-effector pose. To this extent, a soft-region-growing algorithm is presented for effective plane segmentation, even in the case of curved surfaces. An optimization-based quality metric is then used for the evaluation of grasp points to ensure indirect force closure. The proposed grasp framework is compared with the existing state-of-the-art grasp planning approach, Grasp pose detection (GPD), as a baseline over multiple simulated objects. The effectiveness of the proposed approach in comparison to GPD is also evaluated in a real-world setting using image and point-cloud data, with the planned grasps being executed using a ROBOTIQ gripper and UR5 manipulator. The proposed approach shows better performance in terms of higher probability for force closure with complete repeatability. 
\end{abstract}

\section{Introduction}

\IEEEPARstart{G}{rasping}, a fundamental aspect of robotic manipulation, has witnessed considerable attention and advancement in recent years, reflecting the growing importance of robotics in various domains. On the hardware side, grippers have evolved from early two-finger designs to multi-fingered human-like systems. On the grasp planning side, research has progressed to intricate soft grasp planning, necessitating a deeper understanding of object properties, environmental dynamics, and real-time adaptability.



Current research focuses primarily on end-to-end learning-based methods, taking in  RGBD images or occluded point clouds to propose possible grasp poses, which are then executed by the robotic gripper. These approaches try to suggest regions of interest (ROIs) and then try to estimate the confidence by running a classifier network with many cascaded convolution blocks \cite{b6}, \cite{b1}. A fundamental problem with this, as with most learning-based approaches, is the lack of reliability and completeness. In addition to this, such methods require extensive custom-labelled data, often including augmented or synthetic RGBD images to train the networks \cite{b6} \cite{b5}, \cite{b7},\cite{b3}, \cite{b4}. This is often hard to generate and may not work across varied test conditions, as these datasets seldom contain domain randomization, making knowledge transfer between different conditions and test setups difficult. Learning-based methods also take up significant computation needed, especially when running the inference on deep networks; as such, they cannot usually be run on single-board computers.

Another class of grasp planning approach estimates grasp pose by sampling randomly in the six-degree-of-freedom (6-DOF) space around the vicinity of the object \cite{b15} \cite{b39} and then choosing an ideal pose towards maximizing a grasp-quality function. The epsilon ball criterion is often used for multi-fingered hands, while force closure is used for antipodal grasping. These methods overlook a significant implementation issue: the probabilistic nature of the solution. For real-world use cases, it is often vital to have the algorithm propose the same or a similar grasp pose when repeated multiple times with the same input. This is seldom the case with learning and sampling-based approaches. A probabilistic algorithm would output different poses requiring different manipulator trajectories for the same task run at various points in time. 

Grasp Pose Detection (GPD) \cite{b15} is a well-known algorithm for detecting feasible grasp poses for parallel-jaw grippers. The GPD algorithm uses point cloud data to generate candidate grasps by sampling and evaluating multiple 6-DOF grasp poses. One of its key features is using a convolutional neural network (CNN) to assess the quality of the sampled grasps based on the geometric configuration of the object and the contact points. This approach allows GPD to generalize across various object shapes and handle cluttered environments. Its robustness in generating high-quality grasps without requiring object models makes it a widely adopted baseline in robotic grasping research. For the above reasons, GPD will be considered as the baseline for comparison of the novel grasp planning approach presented in this paper. 

\begin{figure*}[t]
    \centering
    \includegraphics[width=1\textwidth]{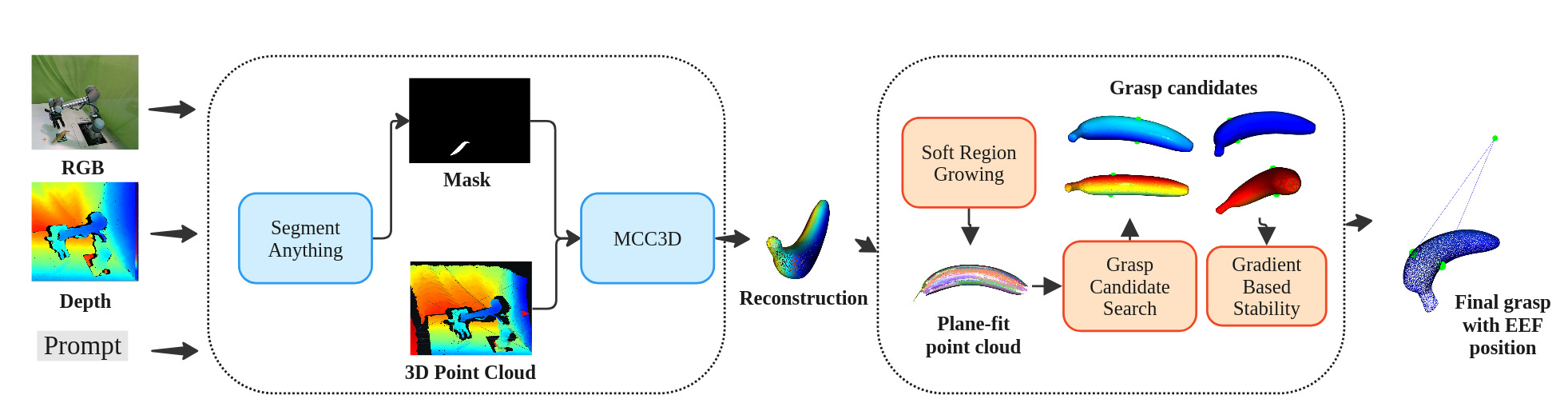}
    \caption{End-to-end real-world grasp pipeline}
    \label{fig:rws}
\end{figure*}

Towards addressing the above-mentioned limitations of existing grasp planning algorithms, we propose QuickGrasp, a novel and lightweight antipodal grasp planning approach. The proposed approach focuses on force closure due to its practical significance in antipodal grasping. To this extent, the proposed approach aims to solve two pivotal and interconnected factors, identify a set of contact points capable of restraining an object by applying suitable forces, and compare two feasible force closure grasps. These questions are inherently intertwined, as there are typically multiple sets of contact points that can ensure force closure. Therefore, this work proposes a unique force-position-based optimization approach to formulate a new quality metric that results in force closure. Starting with a partial point cloud, the proposed approach begins by reconstructing the complete point cloud, using existing algorithms \cite{b35}, \cite{b36},\cite{b37}. A soft-region growing approach is used to segment the point cloud as a collection of planes, which are then used by the optimization algorithm to calculate the optimal grasp. The proposed approach is evaluated over the YCB Objects \cite{b38} in simulation, comparing with the grasps provided by GPD. Finally, the end-to-end grasp planning stack, shown in Fig. \ref{fig:rws}, is evaluated over real-world setups to show its effectiveness in generating candidate grasps for almost any object using a single depth camera.

\section{Methodology}
The proposed approach begins with a soft region-growing algorithm that segments the object's surface into planes, facilitating the identification of candidate grasp regions. These candidates are selected by locating anti-parallel planar patches, which are potential contact points for grasping. Then a force modeling step is performed in which the contact forces are calculated and transformed to ensure force closure at these contact points. An analysis on the stability of the grasp is performed to guarantee robustness, assessing the system’s ability to resist external forces and ensuring stable and reliable grasps throughout the process. The overall workflow is presented in Algorithm \ref{alg:grasping_approach}. Each of the individual step is explained in detail in the following subsections.

\subsection{Soft Region Growing}
\begin{figure}[t]
    \centering
    \includegraphics[width=1\linewidth]{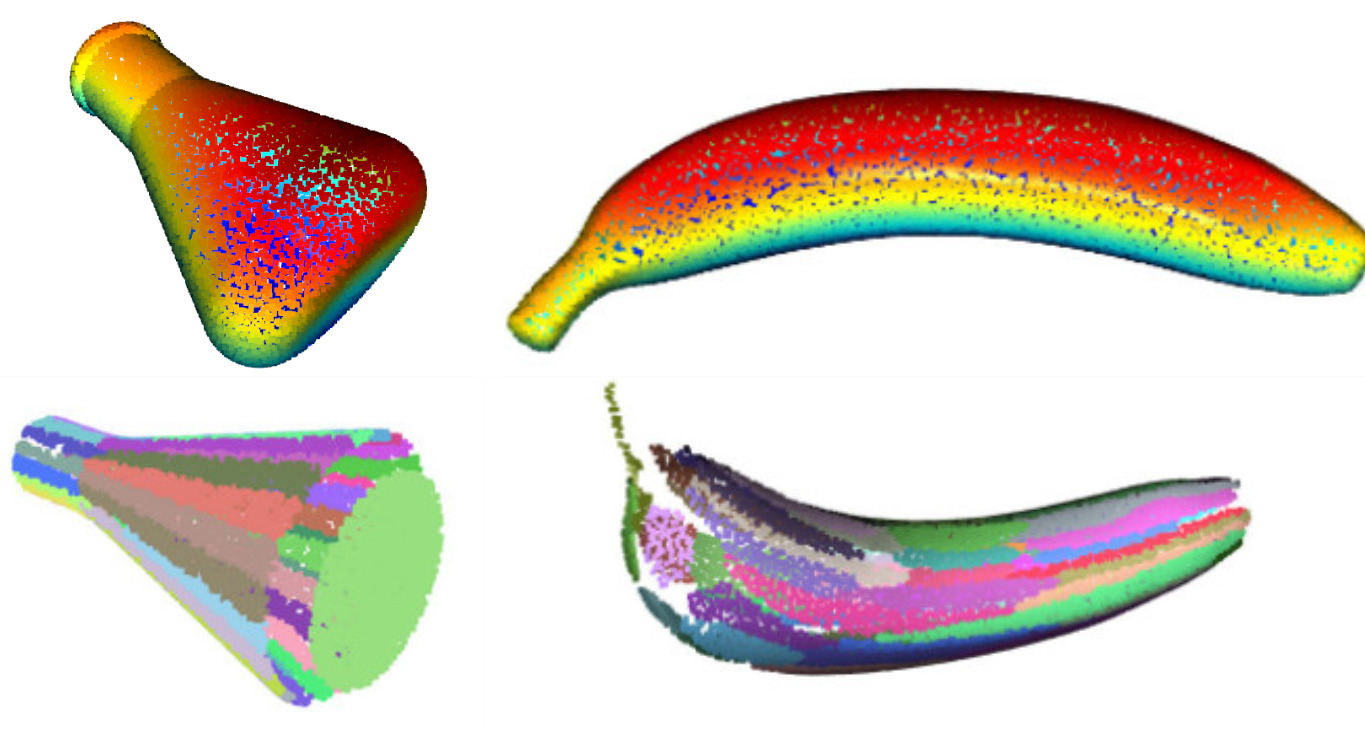}
    \caption{Soft region growing to produce planes on complete point cloud of a banana and an Erlenmeyer Lab flask, the top row depicts the point clouds while the bottom one depicts plane-fit point clouds}
    \label{fig:srg}
\end{figure}

As mentioned in the previous section, the proposed approach begins by decomposing the object's surface into a collection of planes. Numerous algorithms exist for point-cloud plane fitting, such as RANSAC and region growing. The proposed approach uses a variant of the region-growing algorithm to fit planes with tolerances. Region growing was used due to its deterministic iterative growth, which allows for reliable plane fitting on primitive shapes and everyday objects, such as a mug, a lab flask and even a banana as shown in Fig. \ref{fig:srg}. The resulting planes are queried for anti-parallel sets of planes with overlapping patches. Since the original input for the approach was a partial point cloud, the reconstructed regions are given during the querying. A heuristic approach is used, combining reconstruction confidence, viewed as a belief of the point and the Euclidean distance of the point, to choose the grasp points that will be antipodal.

In this work, traditional region growing techniques \cite{b11} are extended into a soft region growing approach with fine-tuned parameters. While classical region growing methods rely on rigid constraints to classify points, the proposed approach integrates tolerance thresholds for distance and angular deviations, making it robust for decomposing curved surfaces as a collection of planes. Given a 3D point cloud, least-squares plane fitting is used as the primary method for region detection. The region-growing process begins by identifying seed points and expands by incorporating neighbouring points based on their proximity and alignment with the local surface. Unlike rigid models, the proposed soft variant allows points slightly deviating from the perfect plane to be still included in the region. This approach is particularly practical for real-world data, where imperfections and noise are expected. It can segment point clouds with complex surfaces, capturing flat planes and subtly curved structures as collections of locally fitted planes. This soft constraint improves robustness and reduces the over-segmentation issues in traditional methods. The proposed soft-region growing builds on top of the \cite{b34} CGAL implementation, as explained below.

\section*{Soft Region Growing}

\textbf{Input:}
\begin{itemize}
    \item Point cloud $\{P\}$, Normals $\{N\}$, Curvatures $\{c\}$
    \item Neighbour function $\Omega(.)$, Curvature threshold $c_{th}$, Angle threshold $\theta_{th}$
\end{itemize}

\textbf{Steps:}
\begin{enumerate}
    \item Initialize empty region list $R$ and available points list $A$ (all points in $P$).
    
    \item \textbf{While} $A$ is not empty:
    \begin{enumerate}
        \item Select point with minimum curvature from $A$ as seed ($P_{min}$).
        \item Create new region $R_c$ starting with seed $P_{min}$.
        \item \textbf{For each} point in the region (starting with $P_{min}$):
        \begin{enumerate}
            \item Find nearest neighbours using $\Omega$.
            \item \textbf{For each} neighbor:
            \begin{itemize}
                \item If in $A$ and normal deviation is below $\theta_{th}$, add to $R_c$.
                \item If curvature is below $c_{th}$, add to seeds.
                \item Remove from $A$.
            \end{itemize}
        \end{enumerate}
        \item Add completed region $R_c$ to the global list $R$.
    \end{enumerate}
    
    \item Return $R$ (list of regions).
\end{enumerate}

\subsection{Candidate Selection from Planar Patches}
After region growing, the next step is identifying the candidates from the planar patches obtained from the soft region growing. To achieve antipodal grasping, sets of anti-parallel planar patches are selected from the set of planar patches. Intuitively, this is the region between the two planar patches where the gripper can make contact. This overlapping region can be estimated through ICP \cite{b40}. Given \(\mathbf{n}_1\) and \(\mathbf{n}_2\) are the normals of the two patches, the common normal can be calculated as the mean. The projection matrix \(P\) is constructed as:

\[
P = I - \mathbf{n}_{\text{com}} \mathbf{n}_{\text{com}}^T
\]

where \(I\) is the identity matrix, and \(\mathbf{n}_{\text{common}} \mathbf{n}_{\text{common}}^T\) is the outer product of \(\mathbf{n}_{\text{common}}\) with itself. The points in the two-point clouds are then projected onto their common plane. The point clouds are cropped to the overlapping bounding box if some overlap exists. The projected points are then transformed back to their original space using:

\[
\mathbf{overlap}_{1,\text{orig}} = P^{-1} \mathbf{overlap}_{1,\text{proj}}^T
\]

\subsection{Contact Force Modeling and Transformation} 
The forces applied at the planned contact points are modelled as point contacts with coulomb friction along the tangent plane.  The forces at the contacts and on the object are represented by a set of coordinate frames attached at each contact location and the object reference point. No slip and point-contact assumptions were made to consider the location of the contact point on the object to be fixed.  Only force vectors lying outside the friction cone would lead to slipping. $f_{\textnormal{c}_i}\epsilon$ $R$ is the magnitude of the force applied by the finger in the
normal direction. The force applied by a contact is modelled 
as a wrench $F_{\textnormal{c}_i}$ applied at the origin of the contact frame, $C_i$. The resulting wrench on the object can be expressed as the product of the Grasp Basis $B_c$ and the contact force $F_{\textnormal{c}}$. The Friction cone at the contact point is denoted by $FC_{\textnormal{c}}$. In the local contact frame, the Z axis is along the normal to the point of contact and the XY  plane is tangential to the surface.

\begin{equation}
F_{c_i}= B_{c_i}*f_{c_i} \quad f_{c_i} \in F C_{c_i},
\end{equation}
\begin{equation}
B_{c_i}=\left[\begin{array}{lll}1 & 0 & 0 \\ 0 & 1 & 0 \\ 0 & 0 & 1 \\ 0 & 0 & 0 \\ 0 & 0 & 0 \\ 0 & 0 & 0\end{array}\right]
\end{equation}
\begin{equation}
F C_{c_i}=\left\{f \in \mathbb{R}^3: \sqrt{f_{c_{i_x}}^2+f_{c_{i_y}}^2} \leq \mu f_{c_{i_z}}, f_{c_{i_z}} \geq 0\right\}.
\end{equation}

Transformations are obtained to convert the forces to the object coordinate frame to determine the effect of the contact forces on the object.   The configuration of the $i^{th}$ contact frame relative to the object frame is given by  $p_{\textnormal{oc}_i}$, $R_{\textnormal{oc}_i}$ \cite{b41}.
\vspace{0.2em}
\begin{equation}
F_{obj_{frame}}=(\operatorname{Ad}_{g_{o c_i}^{-1}})^\top F_{c_i}=\left[\begin{array}{cc}R_{o c_i} & 0 \\ \widehat{p}_{o c_i} R_{o c_i} & R_{o c_i}\end{array}\right] B_{c_i} f_{c_i}
\end{equation}

The matrix $Ad^\top$ is the wrench transformation matrix, which maps contact wrenches to object wrenches. The contact map, $G$, is the linear map between contact forces, represented with respect to $B_{\textnormal{c}_i}$, and the object wrench. Assuming there are $k$ fingers contacting an object, the total wrench on the object is the sum of the object wrenches due to each finger. Since each contact map is linear, the wrenches can be superposed. The net object wrench is then given by
\begin{equation}
F_o=G_1 f_{c_1}+\cdots+G_k f_{c_k}=\left[\begin{array}{lll}G_1 & \cdots & G_k\end{array}\right]\left[\begin{array}{c}f_{c_1} \\ \vdots \\ f_{c_k}\end{array}\right]
\end{equation}
\begin{equation}
\begin{tabular}{llll}
$G=\left[\begin{array}{lll}\operatorname{Ad}_{g_{o c_1}^{-1}}^\top & \cdots & \operatorname{Ad}_{g_{o c_k}^{-1}}^\top B_{c_k}\end{array}\right]$ \\
\
\end{tabular}
\end{equation}
\begin{equation}
F_o=G f_c \quad f_c \in F C,
\end{equation}

The above formulation would give the optimal forces to be applied at the contact regions to ensure force closure.

\subsection{Grasp Stability Analysis}
A good grasp should have force closure, allowing it to resist large external forces, including its weight and other external wrenches. Considering this, various grasp quality measures\cite{b50} have been discussed utilizing methods with the grasp matrix \cite {b17} $G$, geometric constraints \cite {b19, b20, b21,b22,b100}, and limitations considering the finger forces. For antipodal grasping, traditional methods like Grasp wrench space volume and epsilon radius criteria \cite{b41} can sometimes be misleading since the GWS for two-fingered antipodal graspings is degenerate. For example, the best antipodal grasping for a cylinder is at two points situated diametrically opposite on the curved surface and a line passing through the centre of mass (COM). Still, another grasp without force closure and contact points that don't lie along the line passing through the COM could have a higher score under the above-mentioned criteria, even though the first grasp is more human-like.

For two-fingered antipodal grasping, the best quality criterion is force closure itself. Force closure can be checked by analyzing the eigenvalues of the Grasp Matrix $G$ \cite{b33}. Taking the recent success of gradient-based quality criteria, a new criterion is devised to test the stability of the grasp. This criterion considers the friction cone constraints and optimizes the forces to resist perturbations in any direction. This provides an alternative way to think of grasp quality as a measure of stability. This new quality, coupled with the proposed grasp planning and candidate selection, indirectly ensures force closure over 90 per cent of the time in our experiments as shown in Table \ref{tab:comp}. This is formulated as a non-linear constrained optimization problem that could be solved using sequential least squares programming methods. The grasp quality can now be specified as minimising the following objective function for the non-linear optimization problem.

\begin{equation}
\begin{gathered}
    \text{Cost} = \sum_{i=1}^{8}\prod_{j=1}^{3} \left( \mathbf{f}_{c}^\top G^\top G \mathbf{f}_{c} - \mathbf{F}_{\text{ex}_j}^\top \mathbf{F}_{\text{ex}_j} \right) \\
    F C_{c_i} = \left\{ \mathbf{f}_{c} \in \mathbb{R}^3 : \sqrt{f_1^2 + f_2^2} \leq \mu f_3, f_3 \geq 0 \right\}  \\ \text{for } i \in \{1, 2, 3\}
\end{gathered}
\end{equation}

Here, $F_{\textnormal{ex}_j}$ refers to the pseudo force that acts in all the directions at the centre of mass of the object; this is to simulate possible forces acting during motion. These forces are selected to form the basis for each of the eight octants of the real space. The proposed objective function consists of the product of the difference with each positively spanning basis and summing it for all such spanning bases, which is then minimized to obtain grasp points. 
The 6-DOF pose for the end-effector can then be determined based on the chosen contact points and the geometrical dimensions of the gripper; since the contact is a two-fingered contact, a motion planner could be utilized to plane a feasible path which also avoids collisions. The entire methodology is presented in Fig. \ref{fig:rws} in a concise manner, as well as in the form of Algorithm \ref{alg:grasping_approach}.


\begin{algorithm}
\caption{Psuedocode of the Algorithm}
\label{alg:grasping_approach}

\KwIn{\textbf{Input: }Point Cloud \( P \)}
\KwOut{\textbf{Output: }Best Grasp Candidate $GC_{opt}$}

\BlankLine
\textbf{1. Preprocessing:}
\begin{itemize}
    \item Filter noise and downsample \( P \).
\end{itemize}

\BlankLine
\textbf{2. Soft Region Growing:}
\begin{itemize}
    \item Initialize \( \text{SegmentedRegions} \).
    \item \ForEach{point \( p \) in \( P \)}{
        \begin{itemize}
            \item Grow region \( R \) around \( p \) based on distance and angular thresholds.
            \item Append \( R \) to \( \text{SegmentedRegions} \).
        \end{itemize}
    }
\end{itemize}

\BlankLine
\textbf{3. Candidate Selection:}
\begin{itemize}
    \item Initialize \( \text{GraspCandidates} \).
    \item \ForEach{pair \( (R_1, R_2) \) in \( \text{SegmentedRegions} \)}{
        \If{\( R_1 \) and \( R_2 \) are antiparallel and overlapping}{
            Add \( (R_1, R_2) \) to \( \text{GraspCandidates} \).
        }
    }
\end{itemize}

\BlankLine
\textbf{4. Contact Force Modeling:}
\begin{itemize}
    \item \ForEach{candidate \( C \) in \( \text{GraspCandidates} \)}{
        Compute and transform contact forces for \( C \).
    }
\end{itemize}

\BlankLine
\textbf{5. Grasp Stability Analysis:}
\begin{itemize}
    \item \ForEach{candidate \( C \) in \( \text{GraspCandidates} \)}{
        Calculate the stability score using wrench equilibrium and force closure. Return the grasp with the best stability score as $GC_{opt}$
    }
\end{itemize}
\end{algorithm}

\section{Validation}

 The proposed grasp planning approach was validated in simulation and real-life experiments. Performance under the proposed grasp planning approach was compared with Grasp Pose Detection (GPD) as a baseline using the Robust Force Closure Grasp quality metric, as explained in the following subsection. 

\subsection{ Robust Force Closure Grasp Metric}
 The Robust Force Closure grasp metric checks the probability of force closure under perturbations, as introduced in  DexNet\cite{b32}. The metric requires introducing small random perturbations to the end effector pose and then estimates the probability of the resultant perturbed grasp attaining force closure, over 100 trials. Since the proposed planner works on finding points on the object, random noise $\mathcal{N}(0,\sigma^2)$ was introduced on these points to get the closest perturbed point on the surface of the object. force closure was then checked at the perturbed points. Grasp planning using GPD gives the desired 6-DOF pose for the gripper. The contact points under GPD were approximated by projecting the gripper pose onto the object and applying perturbations to the projected points.
 
 Validation over simulation was performed over objects from the YCB dataset \cite{b38}, with different noise parameters to evaluate the grasp proposed by both planners. The robust Force Closure grasp metric was calculated by constructing the grasp matrix and performing singular value decomposition. Given all the singular values are above a certain threshold, the grasp is classified as a force closure grasp. The threshold for all experiments presented in the paper was kept at 0.01. Complete point clouds were given to both algorithms to ensure a fair comparison and evaluate the performance.

\begin{table}[b]

\centering
\caption{Comparison Across Different Perturbations (0.02, 0.05, 0.1)}
\label{tab:comp}
\resizebox{\columnwidth}{!}{%
\begin{tabular}{|l|c|c|c|c|c|c|}
\hline
\textbf{Object} & \multicolumn{2}{c|}{\textbf{Perturbation 0.02}} & \multicolumn{2}{c|}{\textbf{Perturbation 0.05}} & \multicolumn{2}{c|}{\textbf{Perturbation 0.1}} \\
\cline{2-7}
 & \textbf{Ours} & \textbf{GPD} & \textbf{Ours} & \textbf{GPD} & \textbf{Ours} & \textbf{GPD} \\
\hline
Foam Brick       & 1.0  & 0.996 & 1.0  & 0.996 & 0.99 & 0.986 \\
Gelatin Box      & 1.0  & 0.99  & 1.0  & 0.996 & 0.97 & 0.976 \\
Pear             & 1.0  & 0.996 & 1.0  & 1.0   & 1.0  & 0.998 \\
Chips Can        & 0.98 & 0.996 & 1.0  & 1.0   & 0.99 & 0.988 \\
Power Drill      & 1.0  & 0.918 & 1.0  & 0.910 & 1.0  & 0.890 \\
Tennis Ball      & 1.0  & 0.996 & 1.0  & 0.998 & 0.99 & 0.998 \\
Mustard Bottle   & 1.0  & 0.998 & 1.0  & 0.996 & 1.0  & 1.0   \\
Master Chef Can  & 1.0  & 1.0   & 1.0  & 1.0   & 1.0  & 0.996 \\
Tomato Soup Can  & 1.0  & 0.994 & 1.0  & 0.998 & 1.0  & 0.992 \\
Cracker Box      & 0.98 & 0.998 & 0.98 & 0.998 & 0.99 & 0.99  \\
Banana           & 1.0  & 1.0   & 0.98 & 0.996 & 0.99 & 0.986 \\
Hammer           & 0.89 & 0.578 & 0.79 & 0.848 & 0.85 & 0.928 \\
Potted Meat Can  & 1.0  & 0.998 & 1.0  & 0.998 & 1.0  & 0.998 \\
Scissors         & 0.94 & 0.708 & 0.78 & 0.718 & 0.77 & 0.812 \\
Strawberry       & 0.18 & 0.376 & 0.19 & 0.318 & 0.17 & 0.254 \\
Medium Clamp     & -    & 0.676 & -    & 0.896 & -    & 0.896 \\
\hline
\end{tabular}
}
\end{table}
\begin{figure}[t]
    \centering
    \includegraphics[width=0.8\linewidth]{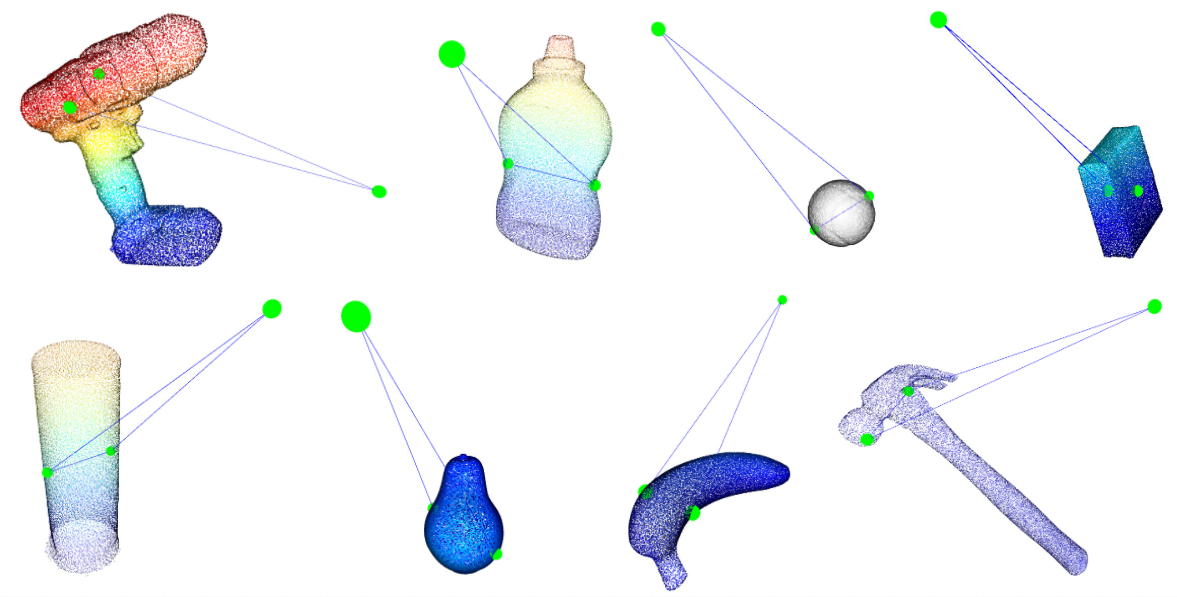}
    \caption{Experimental Results on the YCB dataset, the EEF (outside) is marked in green along with the two gripper contact points }
    \label{fig:sr}
\end{figure}
\subsection{Simulation Results}
Grasp planning results from the proposed approach and GPD under varying perturbations are presented in Table \ref{tab:comp}. In comparison with the proposed approach, GPD shows almost similar or slightly poor performance due to the underlying sampling-based and probabilistic approach. Furthermore, the CNN classifier used to classify grasps within GPD does not extend well to unseen objects without additional training. The above data indicates that some objects, like scissors or strawberries, it seldom finds a grasp that satisfies force closure. The algorithms perform equally well for flat and boxy objects; our algorithm performs slightly better in some cases for objects with complex shapes and high curvatures.

In comparison, our algorithm finds completely antipodal grasps, and with our stability criterion, we get grasps that are always near the COM, which is favourable for reducing the actuator effort.
Some of the proposed poses from our algorithm are shown in Fig. \ref{fig:sr}. One notable aspect in the result is that the proposed approach failed in the case of Medium Clamp, this is likely due to the geometric complexity involved. 

The average time for grasp planning under the proposed approach over the simulated objects was around two seconds on a lightweight laptop with 8GB RAM and 6 cores. In comparison, existing sampling-based planners like Graspit \cite{b14} take about 50s and 50,000 samples to generate a good grasp. This demonstrates that the proposed approach is about 15 times faster, making it a practical choice for real-world applications. The algorithm is also implemented on a Raspberry Pi 4, a single-board computer, achieving a runtime of $~$3.3s, making it suitable for application in mobile setups.

\subsection{Experimental Setup}


\begin{figure}[t]
    \centering
    \includegraphics[width=0.75\linewidth]{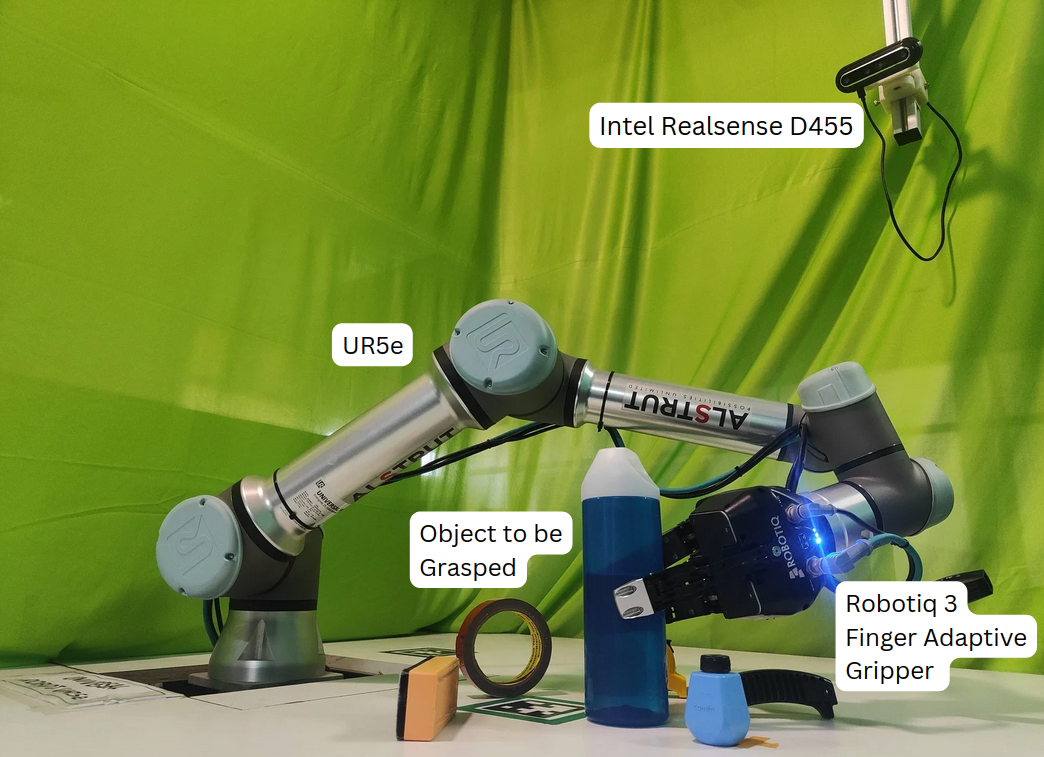}
    \caption{Experimental setup - UR5e manipulator and ROBOTIQ 3-finger adaptive gripper in pinch mode with the Realsense D455 depth camera}
    \label{Fig:Exp_setup}
\end{figure}

The proposed grasp planning approach was validated in real-world conditions as an end-to-end system, shown in Fig. \ref{fig:rws}. The hardware setup is shown in Fig. \ref{Fig:Exp_setup}. The perception-reconstruction-planning-control pipeline used for the experimental validation is explained below. 

A Realsense D455 depth camera was used to extract RGB and depth data from the scene. The image was then used to prompt the Segment Anything Model (SAM) \cite{b37}. A partial point cloud is generated from the segmented masks and the incoming RGBD data. For the point cloud reconstruction, Multi-View Compressive Coding (MCC) was employed with the MCC3D stack \cite{b35}, which provides a full reconstruction with confidence labels for each point. MCC3D was chosen as it efficiently combines RGB image data and point clouds from a single view to achieve compressive coding and reconstruction. Unlike traditional multi-view methods that require multiple 2D perspectives, MCC3D jointly encodes geometric (point cloud) and visual (RGB) information from a single viewpoint. This approach ensures efficient point cloud compression while preserving the scene's visual and structural integrity.

The reconstructed point cloud was then passed to the proposed grasp planning approach to identify the best grasping points. These points can then be used to estimate the desired end-effector pose based on the properties of the gripper being used. As shown in Fig.\ref{Fig:Exp_setup}, a ROBOTIQ 3-finger adaptive gripper was used for grasping. The gripper was used in pinch mode to mimic a 2-fingered parallel-jaw gripper. 

Once the desired end-effector pose is estimated, motion planning for the robotic manipulator can be performed to move the gripper from the current pose to the desired pose while avoiding collisions. for the experimental validations, the gripper was mounted on a Universal Robot UR5e manipulator, with motion planning and control being implemented using ROS and MoveIt.

\subsection{Experimental Results}
Experimental validation was performed on three different objects with increasing grasp complexity, a box, a water bottle, and a 3D-printed ellipsoid with high surface curvature.  Five trials were performed for each object with the end-to-end pipeline as explained above. The final executed grasp, planned end-effector pose and grasp candidates for one trial for each object under the proposed approach, and GPD are shown in Table \ref{tab:images1} and \ref{tab:images2}. The qualitative inferences from the experimental validation are presented below:
\begin{itemize}
    \item Referring to rows 1a, 1b and 1c of Table \ref{tab:images1}, we observe that the grasps planned with our algorithm are completely antipodal and along the line passing through the COM, this intuitively shows better stability and human-like grasping. 
    \item Row 2b in Table \ref{tab:images2} shows the GPD proposed grasp for the Bottle. The proposed contact points are at an angle during grasping, leading to significant undesirable displacement of the bottle. This results in the bottle leaning on its side as the grasp is completed. This undesirable motion could be detrimental in real-world applications. The proposed grasp planning algorithm results are shown in row 1b. As seen from the results, we have minimal displacement during grasping itself. The completely antipodal points lead to force closure and zero net wrenches.
    \item Row 2c in Table \ref{tab:images2} shows the GPD proposed grasp for the box. This applies an undesirable non-zero wrench on the object. The result of this can be seen as the large displacement between the pre-grasp and post-grasp position of the box. The final grasped pose of the object deviates significantly from the starting pose.
    \item The predicted and executed grasp match completely for the proposed approach while they often deviate under GPD. Moreover, the proposed approach is completely repeatable, leading to reliable execution in real-world conditions compared to GPD. GPD results are often not repeatable due to its probabilistic nature.
\end{itemize}

\section{Conclusion and Future Work}

This work presented a simple and deterministic grasp planner that tries to resolve some of the issues limiting the real-world use of existing state-of-the-art grasp planning algorithms. With the proposed QuickGrasp approach quick repetitive grasping can be performed without prior information. The proposed approach was validated over simulation and real-world tests, to provide comparable results in most aspects and better results for some aspects, with GPD as the baseline planner. 

The proposed algorithm can be further developed to optimize for forces while handling soft objects. In such cases, better contact point approximations would be needed to minimize the deformation of the grasped object. This can be achieved by optimizing the external Wrench for a given finger position, which could be easily implemented given the knowledge of the contact points.  Ongoing research aims to release the proposed QuickGrasp approach as a general framework for grasping.


\begin{table}[h]
\centering
\begin{tabular}{|c|c|}
\hline
No & Real-world execution :: Final planned grasp :: Grasp candidates \\ \hline
1a & \includegraphics[width=0.75\linewidth]{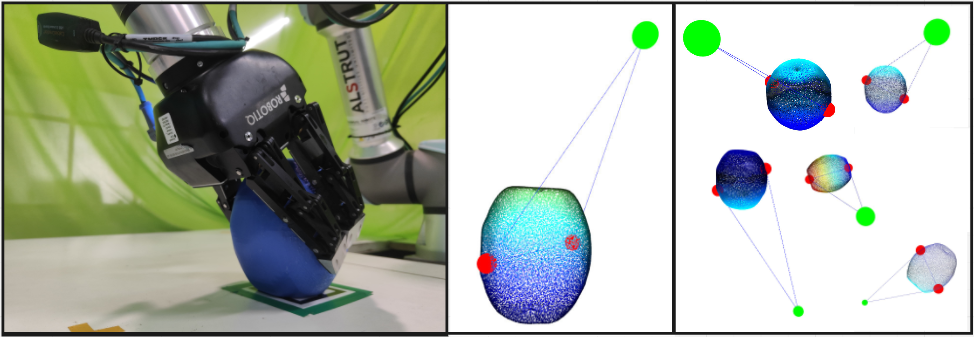} \\ \hline
1b & \includegraphics[width=0.75\linewidth]{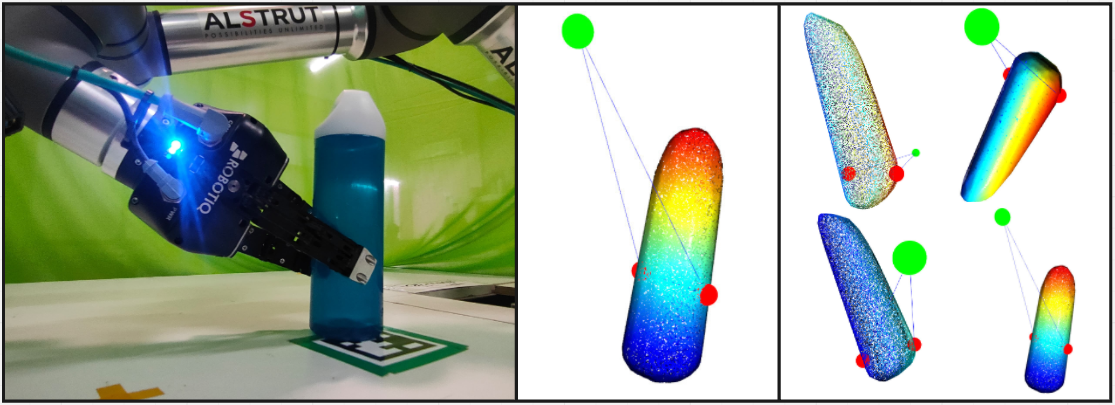} \\ \hline
1c & \includegraphics[width=0.75\linewidth]{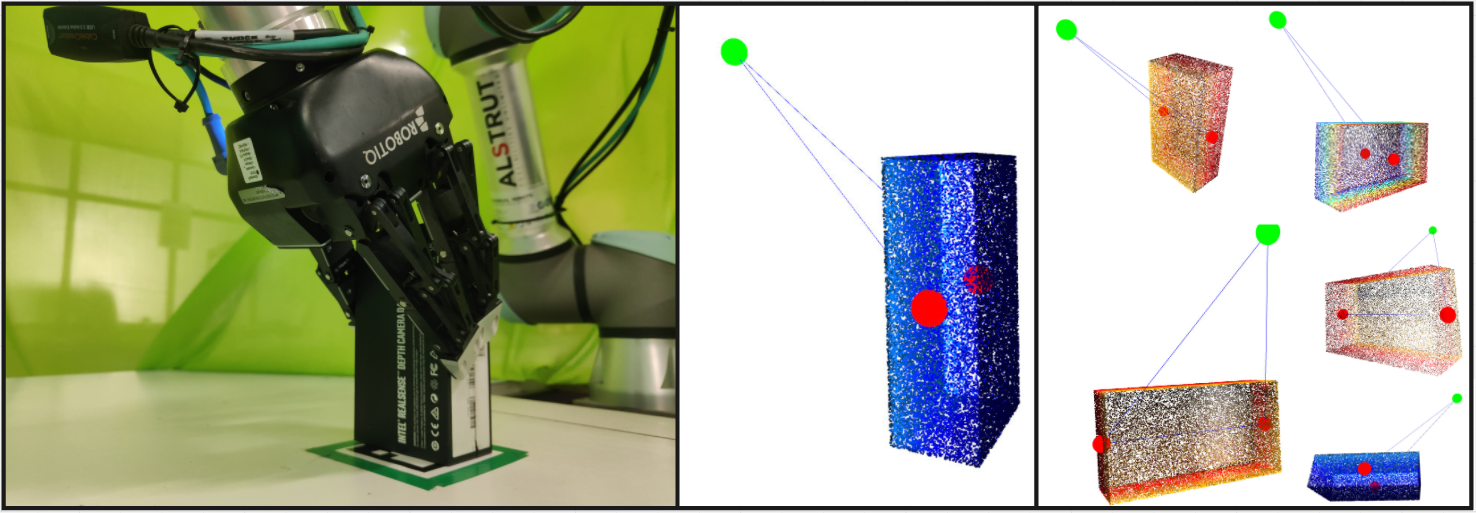} \\ \hline
\end{tabular}
\caption{Experimental results for the grasping of a. Ellipsoid, b. Bottle and c. Box, using the proposed method}
\label{tab:images1}
\end{table}

\begin{table}[h]
\centering
\begin{tabular}{|c|c|}
\hline
No & Real-world execution :: Final planned grasp :: Grasp candidates \\ \hline
2a & \includegraphics[width=0.75\linewidth]{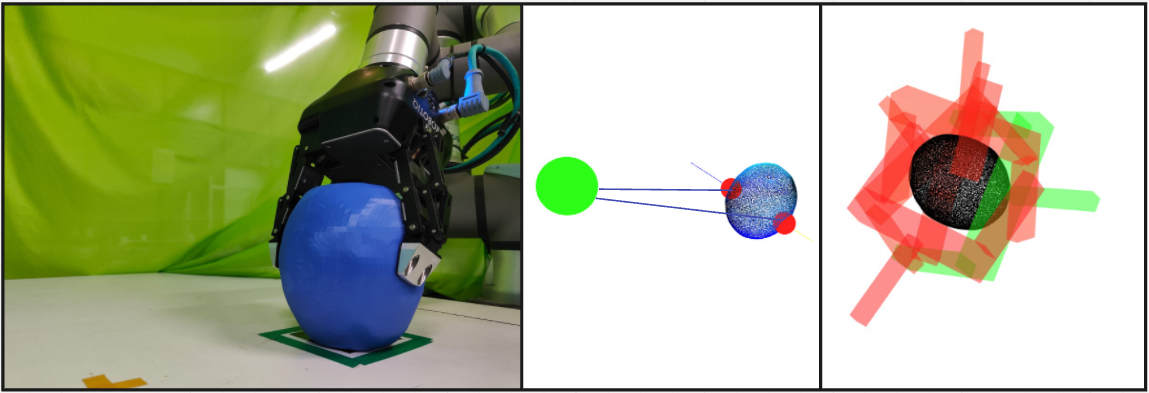} \\ \hline
2b & \includegraphics[width=0.75\linewidth]{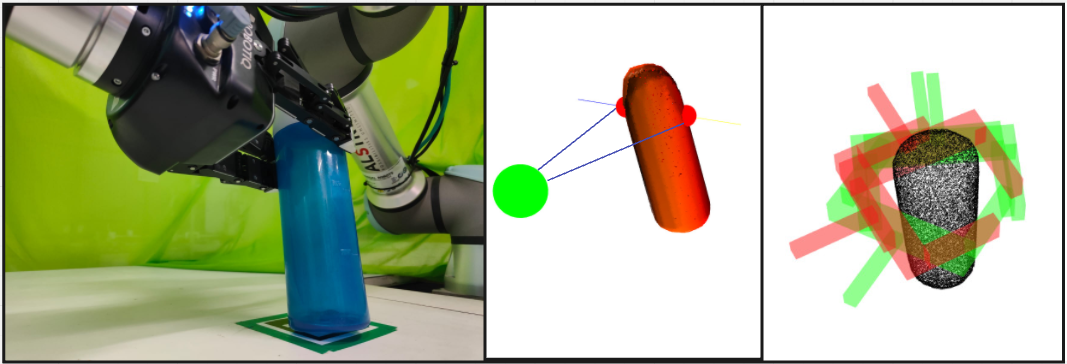} \\ \hline
2c & \includegraphics[width=0.75\linewidth]{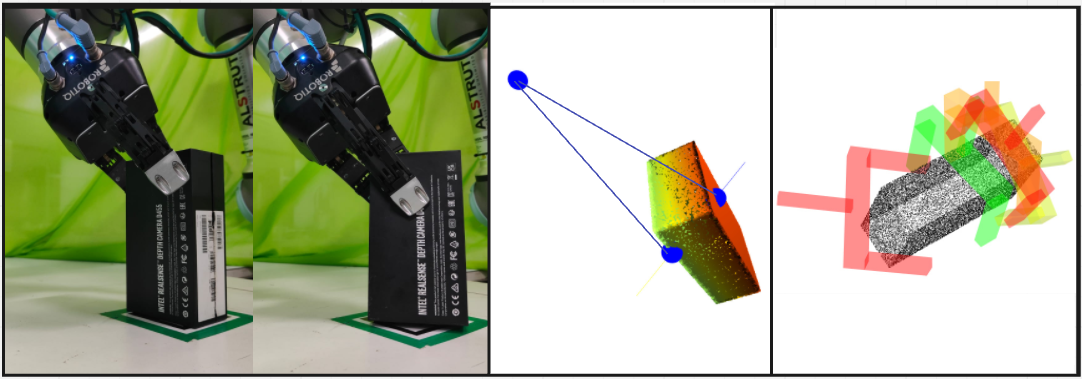} \\ \hline
\end{tabular}
\caption{Experimental results for the grasping of a. Ellipsoid, b. Bottle and b. Box, using GPD }
\label{tab:images2}
\end{table}

\newpage

\vfill

\end{document}